\documentclass[11pt,letterpaper]{article}

\usepackage[T1]{fontenc}
\usepackage[utf8]{inputenc}
\usepackage[margin=1in]{geometry}
\usepackage[hyphens]{url}
\usepackage{graphicx}
\usepackage[round,authoryear]{natbib}
\usepackage{booktabs}
\usepackage{enumitem}
\usepackage{amsmath}
\usepackage{amsfonts}
\usepackage{array}
\usepackage{multirow}
\usepackage[hidelinks]{hyperref}

\hypersetup{
    pdftitle={Rethinking Attention Locality in Spiking Transformers},
    pdfauthor={Zeqi Zheng, Zizheng Zhu, Yuping Yan, Wenxuan Pan, Zhaofei Yu, Yaochu Jin}
}

\title{Rethinking Attention Locality in Spiking Transformers}

\author{
Zeqi Zheng$^{1,2,*}$ \quad
Zizheng Zhu$^{1,*}$ \quad
Yuping Yan$^{2}$ \quad
Wenxuan Pan$^{2}$ \\
Zhaofei Yu$^{3}$ \quad
Yaochu Jin$^{2}$ \\[0.75em]
\small $^{1}$Zhejiang University \quad
$^{2}$Westlake University \quad
$^{3}$Peking University \\
\small $^{*}$Equal contribution
}
\date{}

\begin{document}

\maketitle

\begin{abstract}
Spiking Transformers provide a promising paradigm for efficient visual processing with spike-driven computation, yet their Softmax-free Spiking Self-Attention (SSA) struggles to establish spatially localized token interactions. 
Although existing locality-enhanced SSA methods improve accuracy, it remains unclear whether they consistently induce spatial locality across layers and different Spiking Transformer architectures.
Through Mean Attention Distance (MAD) analysis, we reveal that computational locality does not necessarily translate into spatial locality and show that uniformly applying the same locality enhancement overlooks architecture-dependent deployment requirements.
Motivated by these observations, we propose Spatially Contiguous Local Attention with Boundary Continuity Pathway (SCLA-BCP). 
SCLA computes attention within non-overlapping regions of spatially adjacent tokens, while BCP facilitates cross-boundary information exchange through a lightweight convolutional pathway. 
Furthermore, we develop a hierarchical locality deployment strategy to effectively apply SCLA-BCP across the two major Spiking Transformer architectures.
Extensive experiments on seven static and neuromorphic datasets covering classification, detection, and segmentation demonstrate consistent improvements with limited parameter and energy overhead. 
Notably, our approach improves mAP@50 by up to 9.50\% on COCO 2017 and mIoU by up to 3.42\% on ADE20K. Visualizations, MAD analysis, and ablation studies further validate its effectiveness.
\end{abstract}

\section{Introduction}

Spiking Neural Networks (SNNs), regarded as the third generation of neural networks \citep{maass1997networks, yan2025efficient}, mimic biological neural dynamics and process information through event-driven sparse spikes. 
Unlike Artificial Neural Networks (ANNs) with continuous-valued activations, SNNs use binary spikes and sparse updates, providing advantages in biological plausibility and energy efficiency \citep{indiveri2015memory, eshraghian2023training}. 
In recent years, Transformer architectures \citep{vaswani2017attention, dosovitskiy2020image} have been introduced into SNNs, leading to Spiking Transformers that achieve competitive performance across vision tasks, including image classification \citep{zhou2024qkformer, zhou2026spikingformer}, object detection \citep{miao2025advanced, xu2025hybrid}, and semantic segmentation \citep{lei2025spike2former, yao2025scaling}.

Existing Spiking Transformer backbones can be broadly divided into two architectural families: ViT-like plain architectures \citep{zhou2022spikformer, zhou2026spikingformer} and multi-stage hierarchical architectures \citep{zhou2024qkformer, yao2025scaling}. 
Despite their distinct structures, both families rely on Spiking Self-Attention (SSA) as a key component for token interaction.
To preserve spike-driven computation, SSA removes the Softmax normalization used in conventional attention, leading to relatively uniform attention distributions \citep{qiuquantized, limrethinking} with weak preference for spatially adjacent tokens. 
The resulting insufficient local interactions may constrain the visual representation capability of Spiking Transformers \citep{li2026breaking, zhangneural}.

To strengthen local token interactions, recent studies have introduced explicit spatial locality priors into SSA from two main directions. 
The first modifies the attention computation structure to constrain token interactions. SGLFormer \citep{zhang2024sglformer} introduces Local Spiking Self-Attention (LSSA) through offset-based token grouping, inspiring subsequent locality-oriented designs \citep{liao2025spikeatconv, lee2025spiking}.
The second augments SSA with local receptive fields. 
Localized Receptive Field Spiking Self-Attention (LRF-SSA) \citep{zhangneural} introduces learnable dilated convolution kernels into SSA to impose explicit locality priors on query--key interactions, thereby enhancing neighboring token dependencies.
However, improved task performance alone does not verify whether these mechanisms establish spatially localized attention interactions.
Quantitative evidence of consistently reduced token interaction distances across layers and architectures remains limited.

To investigate this missing evidence, following previous studies \citep{dosovitskiyimage, zhangneural}, we evaluate representative locality-enhanced SSA methods on different Spiking Transformers using Mean Attention Distance (MAD), which quantifies the average spatial range of query--key token interactions.
Our analysis, presented in Section~\ref{sec:problem_analysis}, reveals two findings. 
First, LSSA does not consistently reduce MAD across different backbones.
This can be attributed to its grouping mechanism: as illustrated in Figure~\ref{fig:methods_fig}(a), tokens within the same group are determined by predefined spatial offsets rather than contiguous neighborhoods. 
Therefore, LSSA introduces computational locality but does not necessarily establish spatial locality, exposing a computational--spatial locality discrepancy.
Second, uniformly replacing all SSA layers with LRF-SSA produces layer-dependent effects: MAD remains nearly unchanged or even increases in several layers. 
This observation indicates that applying the same locality-enhancement mechanism to all SSA layers overlooks architecture-dependent locality deployment requirements.

Guided by these findings, we propose Spatially Contiguous Local Attention with Boundary Continuity Pathway (SCLA-BCP) and a hierarchical locality deployment strategy for applying SCLA-BCP to different Spiking Transformer architectures.
SCLA directly addresses the computational--spatial locality discrepancy by partitioning feature maps into non-overlapping, spatially contiguous regions and restricting attention to neighboring tokens within each region.
However, such partitioning limits direct information exchange across region boundaries.
Accordingly, BCP introduces a convolutional pathway mechanism over the original, unpartitioned feature maps and integrates its output through residual fusion.
Under the proposed deployment strategy, SCLA-BCP replaces the first half of the SSA layers in ViT-like plain architectures and is applied to all eligible positions in Stages~1--2 of multi-stage hierarchical architectures, while the remaining network components are left unchanged.
The main contributions are summarized as follows:

\begin{itemize}[label=\textbullet]
    \item We systematically analyze typical locality-enhanced SSA methods and show that their locality mechanisms do not consistently yield spatially closer token interactions, identifying two findings: (i) a discrepancy between computational and spatial locality, and (ii) the potential inadequacy of uniform deployment across all SSA layers.
    \item We propose SCLA-BCP, which aligns the constrained attention scope with spatially contiguous neighborhoods and supports information exchange across local-region boundaries. We further develop a hierarchical locality deployment strategy for effectively applying SCLA-BCP across different Spiking Transformer architectures.
    \item We conduct extensive experiments on representative Spiking Transformer architectures across image classification, object detection, and semantic segmentation. The results show that our method surpasses the corresponding baselines and locality-enhanced SSA variants across the evaluated settings with limited overhead, achieving gains of up to 9.50\% in mAP@50 on COCO 2017 and 3.42\% in mIoU on ADE20K. Evaluations using MAD, attention visualizations, and ablation studies provide further evidence for the proposed design.
\end{itemize}

\section{Related Work}

\subsection{Spiking Transformers}
Existing Spiking Transformer backbones can be broadly categorized into two architectural families:
(a) ViT-like plain architectures tokenize the input through spiking patch embedding and stack multiple Spiking Transformer blocks for feature representation.
Spikformer \citep{zhou2022spikformer} introduced SSA, which used spike-form queries, keys, and values without Softmax normalization.
Spike-driven Transformer (SDT) \citep{yao2023spike} reformulated query--key interactions as spike-domain masking and sparse additions to reduce computational complexity.
Spikingformer \citep{zhou2026spikingformer} further incorporated membrane potential shortcuts to enhance information propagation and representation capability.
(b) Multi-stage hierarchical architectures progressively downsample feature maps to construct multi-scale representations. 
SDT-V2 \citep{yaospike2024} adopted convolution-based and Transformer-based SNN stages to achieve efficient feature extraction across different resolutions.
QKFormer \citep{zhou2024qkformer} proposed spike-form QK attention to reduce attention complexity and enable efficient spiking feature modeling.
Subsequently, SDT-V3 \citep{yao2025scaling} extended SDT-V2 by adopting the spike firing approximation training strategy and more efficient spike-driven convolution and self-attention modules, enabling larger-scale Spiking Transformers.
However, these architectures share a common issue: removing Softmax \citep{bridle1990probabilistic} from SSA to preserve spike-driven computation leads to relatively uniform attention distributions \citep{qiuquantized, limrethinking} and weak interactions among spatially adjacent tokens \citep{zhangneural}, limiting visual representation capability.

\subsection{Locality-Enhanced Spiking Self-Attention}

To strengthen local token interactions, recent studies have incorporated spatial locality priors into SSA through two main directions:
(a) Modifying attention computation structures to constrain token interactions:
SGLFormer \citep{zhang2024sglformer} proposed LSSA, which partitioned feature maps through offset-based token grouping and performed self-attention within each group, thereby restricting the attention computation scope. 
Following the grid attention design of MaxViT \citep{tu2022maxvit}, SpikeAtConv \citep{liao2025spikeatconv} introduced spike-driven grid attention to reorganize spatial token interactions.
(ii) Augmenting SSA with explicit local receptive fields:
LRF-SSA \citep{zhangneural} introduced learnable dilated convolution kernels into the attention matrix to inject locality priors into query--key interactions, thereby encouraging interactions among neighboring tokens.
Subsequently, LSFormer \citep{li2026breaking} proposed spatial dilated attention, which grouped channels and assigned different dilation rates to model multi-scale local attention along horizontal and vertical directions.
Although these methods report performance gains, exploration and analysis of whether they consistently promote spatially localized token interactions across layers and architectures remain limited.

\section{Problem Analysis}
\label{sec:problem_analysis}
This section investigates whether existing locality-enhanced SSA methods consistently induce spatially localized token interactions.
We evaluate two representative methods, LSSA and LRF-SSA, on ImageNet-1K using three different backbones, covering both ViT-like plain and multi-stage hierarchical architectures.
\begin{figure}[htbp]
  \centering
  \includegraphics[width=0.55\linewidth]{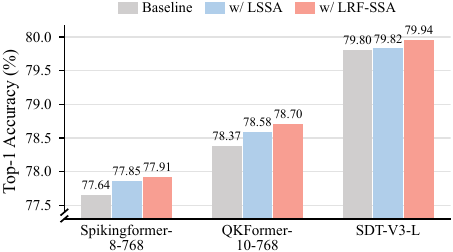}
  \caption{Performance comparison of three Spiking Transformer backbones with LSSA and LRF-SSA trained from scratch on ImageNet-1K. "Baseline" denotes the original backbone without additional modules.}
  \label{fig:problem_analysis_lssa_lrf_ssa}
\end{figure}
As shown in Figure~\ref{fig:problem_analysis_lssa_lrf_ssa}, both methods improve accuracy over their corresponding baselines.
However, performance improvements alone cannot verify whether these methods establish spatial locality.
Therefore, following the previous study \citep{dosovitskiyimage}, we measure MAD (see Appendix~A) to quantify token interaction ranges, where lower MAD values indicate more localized token interactions. 
Our analysis reveals two findings:

\subsection{Computational--Spatial Locality Discrepancy}
\label{sec:computational_spatial_discrepancy}
LSSA constrains token interactions by partitioning feature maps through offset-based token grouping and computing SSA independently within each group.
Given a grouping factor $g$ (default $g=2$), the $(r,c)$-th group, without considering padding, is formulated as
\begin{equation}
\begin{aligned}
\mathcal{G}_{r,c}
=& \{(r+mg,\ c+ng)\mid 0\le m < H/g,\\
&\qquad\; 0\le n < W/g\},
\quad r,c\in\{0,\ldots,g-1\}.
\end{aligned}
\end{equation}
Accordingly, the key-token set for a query at $(y_i,x_i)$ is
\begin{equation}
    \mathcal{N}_i = \{j \mid y_j\equiv y_i\pmod g,\ x_j\equiv x_i\pmod g \}.
\end{equation}
As illustrated in Figure~\ref{fig:methods_fig}(a), tokens within $\mathcal{N}_i$ share identical coordinate residues modulo $g$ but are distributed at regular spatial intervals rather than forming contiguous neighborhoods.
Thus, LSSA reduces the computational scope of attention but does not explicitly enforce interactions among spatially adjacent tokens, revealing a computational--spatial locality discrepancy.
To validate this analysis, following the previous analysis \citep{zhang2024sglformer}, we replace the first SSA layer of each backbone with LSSA, retrain the models on ImageNet-1K, and compare MAD at that layer.  
\begin{figure}[htbp]
  \centering
  \includegraphics[width=0.6\linewidth]{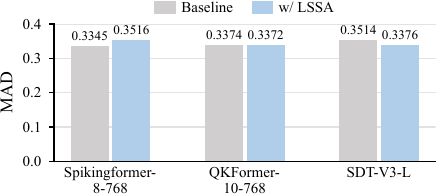}
  \caption{Comparison of MAD between the first SSA layer and its LSSA-replaced counterpart across different Spiking Transformer backbones.}
  \label{fig:problem_analysis_lssa_first_layer}
\end{figure}
Figure~\ref{fig:problem_analysis_lssa_first_layer} shows that LSSA does not consistently reduce these distances across different backbones, with only negligible variations observed. 
This indicates that restricting attention computation through offset-based grouping alone is insufficient to establish spatially localized token interactions, motivating locality mechanisms that explicitly construct regions based on spatial adjacency.

\subsection{Uniform Locality Deployment Issue}
\label{sec:uniform_issue}
LRF-SSA augments SSA with local receptive-field priors and uniformly applies this enhancement to all SSA layers.
However, such uniform deployment assumes that the same locality enhancement is equally effective across all attention layers, which may not consistently induce spatially localized token interactions. 
\begin{figure}[htbp]
  \centering
  \includegraphics[width=0.7\linewidth]{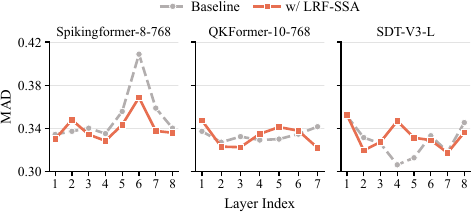}
  \caption{Layer-wise MAD comparison between baseline and LRF-SSA across different Spiking Transformer backbones.}
  \label{fig:problem_analysis_lrf_ssa_all_layer}
\end{figure}
To examine this analysis, following the previous study \citep{zhangneural}, we replace all SSA layers with LRF-SSA in three backbones and compare MAD layer by layer.
As shown in Figure~\ref{fig:problem_analysis_lrf_ssa_all_layer}, uniformly applying LRF-SSA does not consistently reduce MAD across layers, with layer-dependent variations observed across different backbones. 
These results indicate that applying the same locality enhancement to all SSA layers overlooks architecture-dependent locality deployment requirements, motivating a hierarchical locality deployment strategy tailored to different Spiking Transformer architectures.

\section{Method}
\begin{figure}[htbp]
    \centering 
    \includegraphics[width=\linewidth]{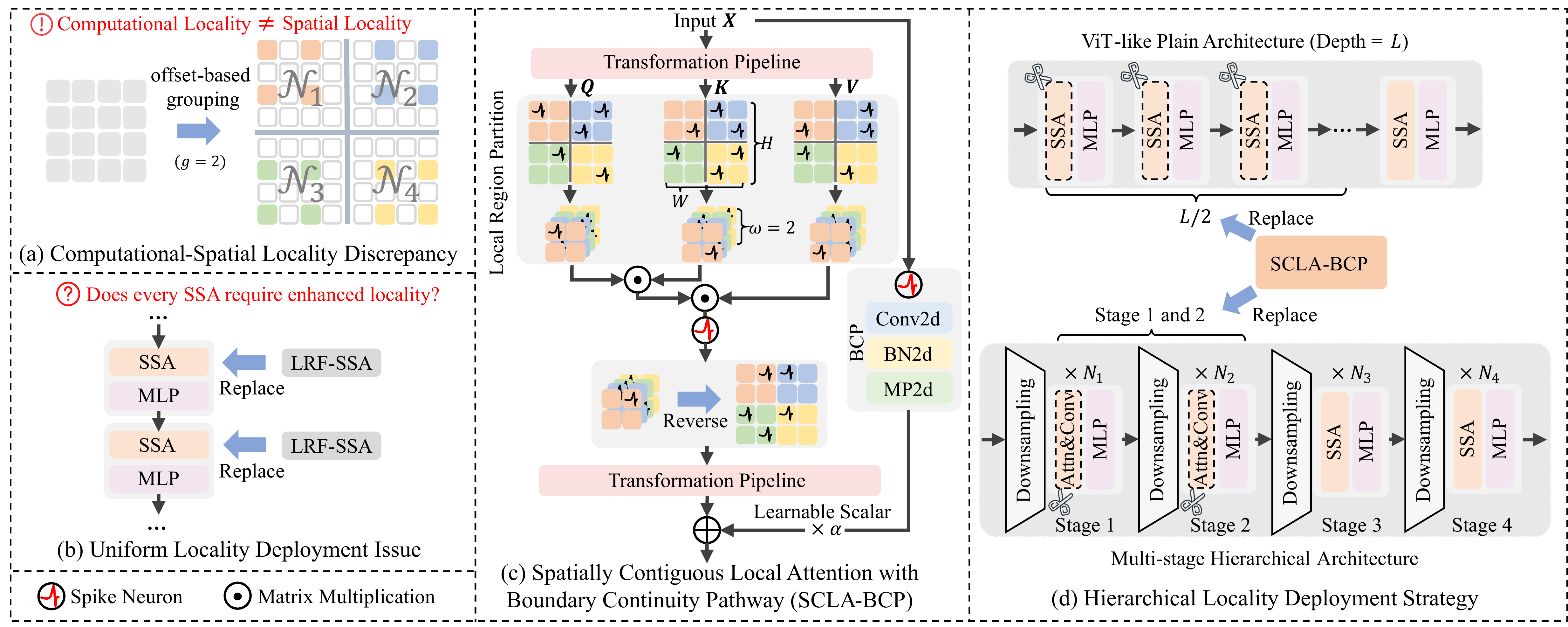}
    \caption{Overview of issues in typical locality-enhanced SSA methods, the proposed SCLA-BCP method, and its hierarchical locality deployment strategy. (a) Computational--spatial locality discrepancy caused by offset-based token grouping in LSSA. (b) Uniform locality deployment issue in LRF-SSA. (c) Pipeline of SCLA-BCP, illustrated with attention computation on a feature map of $H=W=4$ and local region size $w=2$ under a pre-activation backbone for simplicity. (d) Hierarchical locality deployment strategy of SCLA-BCP across different architectures.}
    \label{fig:methods_fig}
\end{figure}

In this section, we introduce Spatially Contiguous Local Attention with Boundary Continuity Pathway (SCLA-BCP) (Figure~\ref{fig:methods_fig}(c)) in Section~\ref{sec:SCLA_BCP} and its corresponding hierarchical locality deployment strategy (Figure~\ref{fig:methods_fig}(d)) in Section~\ref{sec:deployment_strategy}.

\subsection{Spatially Contiguous Local Attention with Boundary Continuity Pathway}
\label{sec:SCLA_BCP}

\subsubsection{Spatially Contiguous Local Attention}
SCLA retains the backbone-specific query, key, and value transformation pipelines while restricting token interactions to non-overlapping, spatially contiguous local regions.
Let $\mathbf{X}^{l}\in\mathbb{R}^{T\times B\times C_l\times H_l\times W_l}$ denote the input to the $l$-th block, where $T$, $B$, $C_l$, $H_l$, and $W_l$ are the number of timesteps, batch size, channels, height, and width, respectively.
Omitting the layer index $l$ for clarity, the transformed features are
\begin{equation}
    \widetilde{\mathbf{U}}_{\xi} =
    \mathcal{G}_{\xi}(\mathbf{X}) =
    \mathcal{S}^{\mathrm{out}}_{\xi}
    \!\left[
    \mathcal{B}_{\xi}
    \!\left(
    \Phi_{\xi}
    \!\left(
    \mathcal{S}^{\mathrm{in}}_{\xi}(\mathbf{X})
    \right)
    \right)
    \right],
\end{equation}
where, for each $\xi\in\{q,k,v\}$, $\mathcal{G}_{\xi}$ denotes the corresponding projection $\Phi_{\xi}$, normalization $\mathcal{B}_{\xi}$, and placement of the leaky integrate-and-fire (LIF) \citep{dayan2005theoretical} activation.
Specifically, $\mathcal{S}^{\mathrm{out}}_{\xi}$ is an identity mapping for pre-activation backbones, whereas $\mathcal{S}^{\mathrm{in}}_{\xi}$ is an identity mapping for post-activation backbones.
The local region size $w$ controls the partition granularity of SCLA.
Following SGLFormer~\citep{zhang2024sglformer}, we set $w=H/2$ for ViT-like plain architectures.
For multi-stage hierarchical architectures, we employ stage-wise decreasing region sizes to match the progressively reduced spatial resolutions across stages, following the prior study~\citep{huang2022orthogonal}.
Detailed configurations are provided in Appendix~C.
Given the selected region size $w$, SCLA first zero-pads the right and bottom boundaries when necessary:
\begin{equation}
    \bar H=\left\lceil\frac{H}{w}\right\rceil w,
    \qquad
    \bar W=\left\lceil\frac{W}{w}\right\rceil w.
\end{equation}
The resulting padded feature map is divided into $M=M_HM_W$ regions, where $M_H=\bar H/w$ and $M_W=\bar W/w$.
Formally, for any $\mathbf{Y}\in\mathbb{R}^{T\times B\times C\times\bar H\times\bar W}$, the local region partition operation $\Pi_w(\cdot)$ is defined by
\begin{equation}
    \left[\Pi_w(\mathbf{Y})\right]_{t,b,m,c,u,v}
    =
    \mathbf{Y}_{t,b,c,iw+u,jw+v},
\end{equation}
where $m=iM_W+j$, $0\le i<M_H$, $0\le j<M_W$, and
$0\le u,v<w$. 
Each region corresponds exactly to one spatially contiguous $w\times w$ block.
Applying $\Pi_w(\cdot)$ to the padded query, key, and value features, splitting the channels into $h$ heads, and flattening each region yields
$\mathbf{Q}_{m,r},\mathbf{K}_{m,r},\mathbf{V}_{m,r}
\in\mathbb{R}^{w^2\times d_r}$ for head $r$ in region $m$.
$d_r$ denotes the head dimension, and the timestep and batch indices are omitted for clarity.
SCLA computes attention independently within each region:
\begin{equation}
\mathbf{A}_{m,r}
=\mathbf{Q}_{m,r}\mathbf{K}_{m,r}^{\top},
\ \
\mathbf{O}_{m,r}
=\mathcal{S}_{\mathrm{LIF}}\!\left(
\mathbf{A}_{m,r}\mathbf{V}_{m,r}
\right),
\end{equation}
where $\mathcal{S}_{\mathrm{LIF}}$ denotes the LIF neuron.
After concatenating the heads, the region-wise features are restored to the padded spatial layout through $\Pi_w^{-1}(\cdot)$ and cropped to the original spatial dimensions.
The cropped feature map is then fed into the backbone-specific output transformation pipeline $\mathcal{G}_o$, producing the SCLA output:
\begin{equation}
\begin{aligned}
\operatorname{SCLA}(\mathbf{X})
={}& \mathcal{G}_o\!\Bigg(
\operatorname{Crop}\!\Bigg[ \Pi_w^{-1}\!\Bigg(
\Big\{\operatorname{Concat}_{r=1}^{h}\mathbf{O}_{m,r}\Big\}_{m=0}^{M-1}
\Bigg)
\Bigg]
\Bigg).
\end{aligned}
\end{equation}

\subsubsection{Boundary Continuity Pathway}
Although SCLA establishes explicit spatial locality, its fixed region boundaries prevent direct interaction between adjacent tokens in different regions.
BCP complements SCLA by processing the intact feature map without region partitioning, thereby enabling information exchange across these boundaries.
For compactness, define
$\mathcal{C}^{l}=\mathcal{P}^{l}\circ\mathcal{B}^{l}\circ\mathcal{D}^{l}$,
where $\mathcal{D}^{l}$, $\mathcal{B}^{l}$, and $\mathcal{P}^{l}$ denote a lightweight $3\times3$ depth-wise convolution, batch normalization, and $3\times3$ maxpooling, respectively.
For an input feature $\mathbf{H}^{l}$, BCP accommodates both pre-activation and post-activation backbones:
\begin{equation}
\operatorname{BCP}^{l}(\mathbf{H}^{l})
=
\begin{cases}
\mathcal{C}^{l}\!\left(\mathcal{S}^{l}(\mathbf{H}^{l})\right),
& \text{pre-activation},\\[1pt]
\mathcal{S}^{l}\!\left(\mathcal{C}^{l}(\mathbf{H}^{l})\right),
& \text{post-activation},
\end{cases}
\end{equation}
where $\mathcal{S}^{l}$ denotes the LIF activation. 
Both $\mathcal{D}^{l}$ and $\mathcal{P}^{l}$ use a stride of 1 and padding of 1, preserving the spatial resolution.
We employ a BCP operating on the block input $\mathbf{X}^{l}$ alongside SCLA.
The complete SCLA-BCP computation is given by
\begin{equation}
\mathbf{\hat{X}}^{l}
=
\mathbf{X}^{l}
+\operatorname{SCLA}^{l}(\mathbf{X}^{l})
+\alpha^{l}\operatorname{BCP}^{l}(\mathbf{X}^{l}).
\end{equation}
The learnable scalar $\alpha^l$ is initialized to $0.1$, limiting perturbations to residual updates early in training while allowing the model to learn the required boundary-compensation strength.
The detailed analysis of the initialization value is presented in Appendix~E.
Following prior reparameterization strategies \citep{yaospike2024, yao2025scaling}, these input-independent scalars can be absorbed into adjacent normalization parameters or synaptic weights, preserving spike-driven computation.

\subsection{Hierarchical Locality Deployment Strategy}
\label{sec:deployment_strategy}

Based on the analysis in Section~\ref{sec:uniform_issue}, we treat deployment depth as a design variable and define distinct SCLA-BCP placement rules for ViT-like plain and multi-stage hierarchical architectures.
Let the $L$ replaceable SSA layers of a backbone be indexed by depth as $l=1,\ldots,L$, and let $K$ denote the deployment depth.
SCLA-BCP is applied to the first $K$ SSA layers. The remaining layers retain their original structures.
According to the deployment study in Section~\ref{sec:SCLA_BCP_deployment}, we use the following architecture-specific settings.

\noindent\textbf{ViT-like plain architectures.}
We set $K=\lfloor L/2\rfloor$, corresponding to replacing the first half of the SSA layers.

\noindent\textbf{Multi-stage hierarchical architectures.}
We set $K$ to the number of replaceable layers in Stages~1 and~2.
Stages~3 and~4 retain their original block structures.

\section{Experiments}
This section evaluates SCLA-BCP and its hierarchical locality deployment strategy through three research questions. 
\textbf{RQ1: How Should SCLA-BCP Be Deployed across Different Spiking Transformer Architectures?} For ViT-like plain and multi-stage hierarchical architectures, what is the optimal deployment depth for each? 
\textbf{RQ2: Effectiveness and Applicability.} Can our proposed method consistently improve performance across different architectures and diverse vision tasks?
\textbf{RQ3: Spatial Locality of Token Interactions.} Does our method enforce localized token interactions as intended and learn more effective spatial representations?

\subsection{Experimental Setup}
We evaluate the proposed approach on seven static and neuromorphic datasets across classification, detection, and segmentation tasks, including CIFAR-10/100 \citep{krizhevsky2009learning}, ImageNet-1K \citep{deng2009imagenet}, CIFAR10-DVS \citep{li2017cifar10}, N-Caltech101 \citep{fei2004learning}, COCO 2017 \citep{lin2014microsoft}, and ADE20K \citep{zhou2017scene}. 
Detailed dataset descriptions and evaluation metrics are provided in Appendix~B.
Experiments are conducted on ViT-like plain (Spikformer \citep{zhou2022spikformer}, SDT \citep{yao2023spike}, and Spikingformer \citep{zhou2026spikingformer}) and multi-stage hierarchical (QKFormer\footnote{All QKFormer-related experiments use the corrected GitHub implementation with identified issues fixed.} \citep{zhou2024qkformer} and SDT-V3 \citep{yao2025scaling}) architectures following their official implementations.
We compare the original models (referred to as baselines hereafter) with LSSA \citep{zhang2024sglformer}, LRF-SSA\footnote{All LRF-SSA-related experiments use the corrected publicly available implementation with ternary spike activation removed.} \citep{zhangneural}, and our method. 
Detailed training configurations are provided in Appendix~C.

\subsection{Hierarchical Locality Deployment Strategy of SCLA-BCP}
\label{sec:SCLA_BCP_deployment}

\begin{figure}[htbp]
  \centering
  \includegraphics[width=0.55\linewidth]{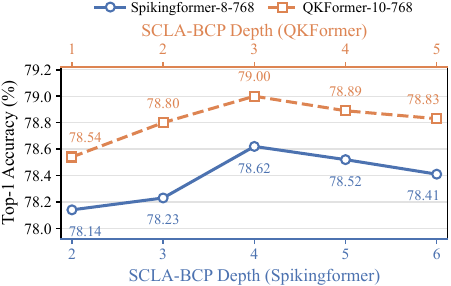}
  \caption{Deployment depth analysis of SCLA-BCP on ImageNet-1K across Spiking Transformer architectures. All deployment variants outperform their backbone baselines (78.37\% for QKFormer and 77.64\% for Spikingformer).}
  \label{fig:exp_deployment}
\end{figure}

To answer \textbf{RQ1}, we investigate the progressive deployment of SCLA-BCP across network depth in ViT-like plain and multi-stage hierarchical architectures. We train all variants from scratch on ImageNet-1K and report their top-1 accuracy. As shown in Figure~\ref{fig:exp_deployment}, QKFormer achieves 79.00\% at depth 3 (all deployable layers in Stages~1--2), while depths 4 and 5 reduce accuracy to 78.89\% and 78.83\%. Spikingformer reaches 78.62\% at depth 4 (the first half of SSA layers), while depths 5 and 6 reduce accuracy to 78.52\% and 78.41\%. These results reveal architecture-specific deployment ranges. Unless otherwise specified, SCLA-BCP is deployed in early layers: the first half of SSA layers for ViT-like plain architectures and the first two stages for multi-stage hierarchical architectures.

\subsection{Effectiveness and Applicability}
To answer \textbf{RQ2}, we evaluate our method on Spiking Transformer backbones from two architectural families across seven static and neuromorphic datasets, covering classification, detection, and segmentation tasks. We compare it with baselines and their LSSA and LRF-SSA variants.

\begin{table}[htbp]
  \centering
  \footnotesize
  \setlength{\tabcolsep}{7pt}
  \renewcommand{\arraystretch}{1.1}

    \begin{tabular}{@{}lccc@{}}
      \toprule
      \multirow{2.5}{*}{\textbf{Backbone}}
        & \multicolumn{1}{c}{\multirow{2.5}{*}{\textbf{Param (M)}}}
        & \multicolumn{2}{c}{\textbf{Top-1 Acc. (\%)}} \\
      \cmidrule(lr){3-4}
        & & \multicolumn{1}{c}{\textbf{CIFAR-10}}
            & \multicolumn{1}{c}{\textbf{CIFAR-100}} \\
      \midrule
      QKFormer
        & 6.74
        & 96.15
        & 80.84 \\
      \quad $+$ LSSA
        & 7.94
        & 96.31
        & 81.07 \\
      \quad $+$ LRF-SSA
        & 6.76
        & 96.43
        & 81.11 \\
      \quad \textbf{$+$ Ours}
        & 6.76
        & \textbf{96.58}
        & \textbf{81.21} \\
      \midrule
      Spikformer
        & 9.32
        & 95.19
        & 77.86 \\
      \quad $+$ LSSA
        & 10.56
        & 95.23
        & 78.34 \\
      \quad $+$ LRF-SSA
        & 9.38
        & 95.37
        & 78.53 \\
      \quad \textbf{$+$ Ours}
        & 9.36
        & \textbf{95.75}
        & \textbf{79.76} \\
      \midrule
      SDT
        & 10.28
        & 95.60
        & 78.40 \\
      \quad $+$ LSSA
        & 12.40
        & 95.54
        & 78.80 \\
      \quad $+$ LRF-SSA
        & 10.29
        & 95.68
        & 79.30 \\
      \quad \textbf{$+$ Ours}
        & 10.30
        & \textbf{95.91}
        & \textbf{79.41} \\
      \midrule
      Spikingformer
        & 9.32
        & 95.95
        & 80.37 \\
      \quad $+$ LSSA
        & 10.56
        & 96.04
        & 80.41 \\
      \quad $+$ LRF-SSA
        & 9.37
        & 95.78
        & 80.52 \\
      \quad \textbf{$+$ Ours}
        & 9.36
        & \textbf{96.23}
        & \textbf{80.91} \\
      \bottomrule
    \end{tabular}%

  \caption{Performance comparison on static datasets with $T=4$. The best results are highlighted in bold.}
  \label{tab:cifar10_cifar100}
\end{table}

\begin{table}[htbp]
\centering
\small
\renewcommand{\arraystretch}{1.1}
\setlength{\tabcolsep}{8pt}

\begin{tabular}{@{}lcrr@{}}
\toprule
\multirow{2}{*}{\textbf{Backbone}}
& \multicolumn{1}{c}{\multirow{2}{*}{\textbf{Param (M)}}}
& \multicolumn{2}{c}{\textbf{Top-1 Acc. (\%)}} \\
\cmidrule(lr){3-4}
& & \multicolumn{1}{c}{$T=10$}
  & \multicolumn{1}{c}{$T=16$} \\
\midrule

QKFormer
    & 1.50 / 2.02
    & 80.30 / 84.69
    & 81.10 / 86.51 \\
\quad $+$ LSSA
    & 2.04 / 2.56
    & 80.70 / 85.18
    & 82.20 / 86.83 \\
\quad $+$ LRF-SSA
    & 1.51 / 2.03
    & 81.30 / 84.93
    & 82.40 / 86.98 \\
\quad \textbf{$+$ Ours}
    & 1.52 / 2.04
    & \textbf{82.80 / 86.99}
    & \textbf{83.50 / 88.34} \\

\midrule

Spikformer
    & 2.57 / 2.60
    & 78.90 / 81.17
    & 80.90 / 83.60 \\
\quad $+$ LSSA
    & 3.12 / 3.15
    & 79.20 / 81.94
    & 81.40 / 83.74 \\
\quad $+$ LRF-SSA
    & 2.60 / 2.62
    & 79.60 / 82.89
    & 82.10 / 83.91 \\
\quad \textbf{$+$ Ours}
    & 2.60 / 2.62
    & \textbf{81.00 / 85.66}
    & \textbf{83.80 / 86.39} \\

\midrule

Spikingformer
    & 2.57 / 2.59
    & 80.20 / 82.88
    & 81.40 / 83.59 \\
\quad $+$ LSSA
    & 3.11 / 3.13
    & 80.60 / 82.39
    & 81.90 / 83.57 \\
\quad $+$ LRF-SSA
    & 2.58 / 2.60
    & 81.70 / 83.11
    & 82.60 / 84.41 \\
\quad \textbf{$+$ Ours}
    & 2.58 / 2.60
    & \textbf{83.10 / 85.91}
    & \textbf{83.60 / 86.76} \\

\bottomrule
\end{tabular}%

\caption{Performance comparison on  neuromorphic datasets. Results are reported in the format of CIFAR10-DVS / N-Caltech101, with the best results highlighted in bold.}
\label{tab:neuromorphic_classification}
\end{table}

\subsubsection{Classification on small-scale static and neuromorphic datasets}

On CIFAR-10 and CIFAR-100, SCLA-BCP improves all four backbones (Table~\ref{tab:cifar10_cifar100}), achieving gains of 0.28--1.90\% over baselines and an average improvement of 0.35\% over the better-performing LSSA or LRF-SSA variant, while introducing only 0.02--0.04M parameters. 
On CIFAR10-DVS and N-Caltech101, it improves all three backbones across timestep settings (Table~\ref{tab:neuromorphic_classification}), achieving 1.83--4.49\% gains over baselines and an average improvement of 1.81\% over the better-performing LSSA or LRF-SSA variant, with only 0.01--0.03M additional parameters.

\begin{table}[htbp]
  \centering
  \footnotesize
  \setlength{\tabcolsep}{5pt}
  \renewcommand{\arraystretch}{1.20}
  \begin{tabular}{@{}lccccc@{}}
    \toprule
    \multirow[c]{2}{*}{\textbf{Backbone}} & \multirow[c]{2}{*}{\textbf{Model Type}} & \textbf{Param} & \textbf{Power} & \textbf{Time} & \textbf{Top-1} \\
                                 &                                      & \textbf{(M)}   & \textbf{(mJ)}  & \textbf{Step} & \textbf{Acc (\%)} \\
    \midrule
    Spikingformer        & Spikingformer-8-768 & 66.36 & 16.95 & \multirow[c]{4}{*}{$4$}          & 77.64 \\
    \quad $+$LSSA              & Spikingformer-8-768 & 71.12 & 18.59 &                               & 77.85 \\
    \quad $+$LRF-SSA           & Spikingformer-8-768 & 66.54 & 18.35 &                               & 77.91 \\
    \quad \textbf{$+$Ours}   & Spikingformer-8-768 & 66.39 & 17.20 &                               & $\mathbf{78.62}_{({+0.98})}$ \\
    \midrule
    QKFormer             & HST-10-768          & 64.96 & 41.65 & \multirow[c]{4}{*}{$4$}          & 78.37 \\
    \quad $+$LSSA              & HST-10-768          & 69.73 & 41.97 &                               & 78.58 \\
    \quad $+$LRF-SSA           & HST-10-768          & 65.12 & 44.24 &                               & 78.70 \\
    \quad \textbf{$+$Ours}   & HST-10-768          & 65.31 & 42.58 &                               & $\mathbf{79.00}_{({+0.63})}$ \\
    \midrule
    SDT-V3               & E-SpikeFormer-S      &  5.10 &  1.70 & \multirow[c]{4}{*}{$1\times4$}  & 75.30 \\
    \quad $+$LSSA              & E-SpikeFormer-S      &  5.11 &  1.73 &                               & 75.36 \\
    \quad $+$LRF-SSA           & E-SpikeFormer-S      &  5.24 &  1.79 &                               & 75.94 \\
    \quad \textbf{$+$Ours}   & E-SpikeFormer-S      &  5.32 &  1.94 &                               & $\mathbf{76.03}_{({+0.73})}$ \\
    \cmidrule(lr){1-6}
    SDT-V3               & E-SpikeFormer-M      & 10.00 &  3.00 & \multirow[c]{4}{*}{$1\times4$}  & 78.50 \\
    \quad $+$LSSA              & E-SpikeFormer-M      & 10.05 &  3.14 &                               & 78.54 \\
    \quad $+$LRF-SSA           & E-SpikeFormer-M      & 10.21 &  3.24 &                               & 78.67 \\
    \quad \textbf{$+$Ours}   & E-SpikeFormer-M      & 10.36 &  3.48 &                               & $\mathbf{79.10}_{({+0.60})}$ \\
    \cmidrule(lr){1-6}
    SDT-V3               & E-SpikeFormer-L      & 19.00 &  5.90 & \multirow[c]{4}{*}{$1\times4$}  & 79.80 \\
    \quad $+$LSSA              & E-SpikeFormer-L      & 19.14 &  5.92 &                               & 79.82 \\
    \quad $+$LRF-SSA           & E-SpikeFormer-L      & 19.25 &  5.91 &                               & 79.94 \\
    \quad \textbf{$+$Ours}   & E-SpikeFormer-L      & 19.48 &  5.99 &                               & $\mathbf{80.31}_{({+0.51})}$ \\
    \bottomrule
  \end{tabular}%

  \caption{Performance comparison on ImageNet-1K with $224 \times 224$ inputs. Best results are highlighted in bold. Energy consumption follows previous works \citep{horowitz20141, chowdhury2022towards} (details in Appendix~D).}
  \label{tab:imagenet1k_comparison}
\end{table}

\subsubsection{Classification on ImageNet-1K}

On ImageNet-1K (Table~\ref{tab:imagenet1k_comparison}), our method consistently improves Spikingformer, QKFormer, and SDT-V3, achieving 0.51--0.98\% gains over baselines (0.69\% on average) and outperforming LSSA and LRF-SSA by 0.58\% and 0.38\% on average, respectively. 
Notably, on Spikingformer, it raises accuracy from 77.64\% to 78.62\%, exceeding LSSA and LRF-SSA by 0.77\% and 0.71\%, respectively.
Our approach introduces only 0.03--0.48M parameters and 0.09--0.93 mJ energy overhead over baselines, while requiring at most 0.23M parameters and 0.24 mJ additional energy over LRF-SSA variants.

\subsubsection{Object Detection and Semantic Segmentation}

We further evaluate our method on dense prediction with COCO 2017 and ADE20K. On COCO 2017 (Table~\ref{tab:coco17_comparison}), our approach improves SDT-V3-S/M by 9.50\%/5.60\% mAP@50 over baselines and surpasses the stronger locality-enhanced variant (LRF-SSA) by 3.70\%/3.60\%. It introduces only 0.21M/0.34M parameters and 0.53mJ/0.28mJ energy overhead over baselines, and 0.07M/0.14M parameters and 0.26mJ/0.12mJ over LRF-SSA.
\begin{table}[htbp]
  \centering
  \small
  \setlength{\tabcolsep}{3pt}
  \renewcommand{\arraystretch}{1.20}
  \begin{tabular}{@{}lccccc@{}}
    \toprule
    \multirow[c]{2}{*}{\textbf{Backbone}} & \multirow[c]{2}{*}{\textbf{Model Type}} & \textbf{Param} & \textbf{Power} & \textbf{Time} & \textbf{mAP@50} \\
                                 &                                      & \textbf{(M)}   & \textbf{(mJ)}  & \textbf{Step} & \textbf{(\%)} \\
    \midrule
    SDT-V3                       & E-SpikeFormer-S & 10.05 & 34.54 & \multirow[c]{4}{*}{$1 \times 4$} & 34.70 \\
    \quad $+$LSSA                      & E-SpikeFormer-S & 10.15 & 34.67 &                              & 39.30 \\
    \quad $+$LRF-SSA                   & E-SpikeFormer-S & 10.19 & 34.81 &                              & 40.50 \\
    \quad \textbf{$+$Ours} & E-SpikeFormer-S & 10.26 & 35.07 &                              & $\mathbf{44.20}_{({+9.50})}$ \\
    \cmidrule(lr){1-6}
    SDT-V3                       & E-SpikeFormer-M & 21.17 & 61.68 & \multirow[c]{4}{*}{$1 \times 4$} & 46.20 \\
    \quad $+$LSSA                      & E-SpikeFormer-M & 21.27 & 61.70 &                              & 47.40 \\
    \quad $+$LRF-SSA                   & E-SpikeFormer-M & 21.37 & 61.84 &                              & 48.20 \\
    \quad \textbf{$+$Ours} & E-SpikeFormer-M & 21.51 & 61.96 &                              & $\mathbf{51.80}_{({+5.60})}$ \\
    \bottomrule
  \end{tabular}%

  \caption{Performance comparison on COCO 2017. Best results are highlighted in bold. Detailed energy consumption is reported in Appendix~D.}
  \label{tab:coco17_comparison}
\end{table}
\begin{table}[htbp]
  \centering
  \small
  \setlength{\tabcolsep}{3pt}
  \renewcommand{\arraystretch}{1.08}
  \begin{tabular}{@{}lccccc@{}}
    \toprule
    \multirow[c]{2}{*}{\textbf{Backbone}} & \multirow[c]{2}{*}{\textbf{Model Type}} & \textbf{Param} & \textbf{Power} & \textbf{Time} & \textbf{mIoU} \\
                                                 &                                      & \textbf{(M)}   & \textbf{(mJ)}  & \textbf{Step} & \textbf{(\%)} \\
    \midrule
    SDT-V3                       & E-SpikeFormer-S &  6.60 & 17.94 & \multirow[c]{4}{*}{$1 \times 4$} & 37.74 \\
    \quad $+$LSSA                      & E-SpikeFormer-S &  6.63 & 18.31 &                         & 38.20 \\
    \quad $+$LRF-SSA                   & E-SpikeFormer-S &  6.73 & 18.13 &                         & 38.51 \\
    \quad \textbf{$+$Ours} & E-SpikeFormer-S &  6.81 & 18.79 &                         & $\mathbf{41.16}_{({+3.42})}$ \\
    \cmidrule(lr){1-6}
    SDT-V3                       & E-SpikeFormer-M & 11.49 & 27.60 & \multirow[c]{4}{*}{$1 \times 4$} & 40.10 \\
    \quad $+$LSSA                      & E-SpikeFormer-M & 11.56 & 28.45 &                         & 40.50 \\
    \quad $+$LRF-SSA                   & E-SpikeFormer-M & 11.67 & 27.78 &                         & 40.69 \\
    \quad \textbf{$+$Ours} & E-SpikeFormer-M & 11.82 & 27.89 &                         & $\mathbf{42.33}_{({+2.23})}$ \\
    \cmidrule(lr){1-6}
    SDT-V3                       & E-SpikeFormer-L & 20.35 & 34.74 & \multirow[c]{4}{*}{$1 \times 4$} & 40.73 \\
    \quad $+$LSSA                      & E-SpikeFormer-L & 20.23 & 36.82 &                         & 41.60 \\
    \quad $+$LRF-SSA                   & E-SpikeFormer-L & 20.62 & 34.98 &                         & 41.70 \\
    \quad \textbf{$+$Ours} & E-SpikeFormer-L & 20.85 & 35.44 &                         & $\mathbf{43.72}_{({+2.99})}$ \\
    \bottomrule
  \end{tabular}%

  \caption{Performance comparison on ADE20K. Best results are highlighted in bold. Detailed energy consumption is reported in Appendix~D.}
  \label{tab:ade20k_comparison}
\end{table}
On ADE20K (Table~\ref{tab:ade20k_comparison}), our method achieves 41.16\%, 42.33\%, and 43.72\% mIoU on SDT-V3-S/M/L, improving over baselines by 3.42\%/2.23\%/2.99\% and over the stronger LRF-SSA variants by 2.65\%/1.64\%/2.02\%. On average, it introduces only 0.35M parameters and 0.61mJ energy overhead over baselines, and 0.15M parameters and 0.41mJ over LRF-SSA.

\subsection{Spatial Locality of Token Interactions}
\begin{figure}[htbp]
  \centering
  \includegraphics[width=\linewidth]{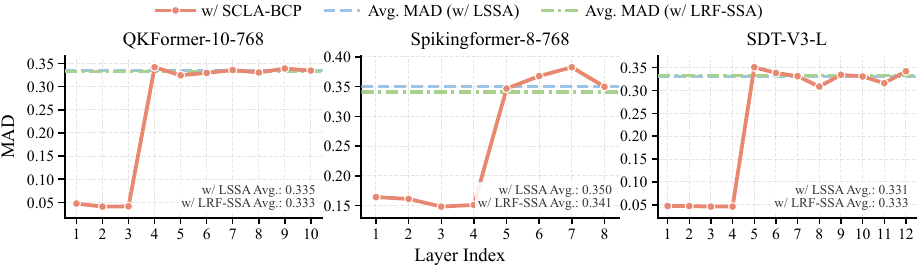}
  \caption{Layer-wise MAD of SCLA-BCP compared with the average MAD of LSSA and LRF-SSA on ImageNet-1K.}
  \label{fig:exp_SCLA_BCP_compare_variants}
\end{figure}

\begin{figure}[htbp]
  \centering
  \includegraphics[width=0.9\linewidth]{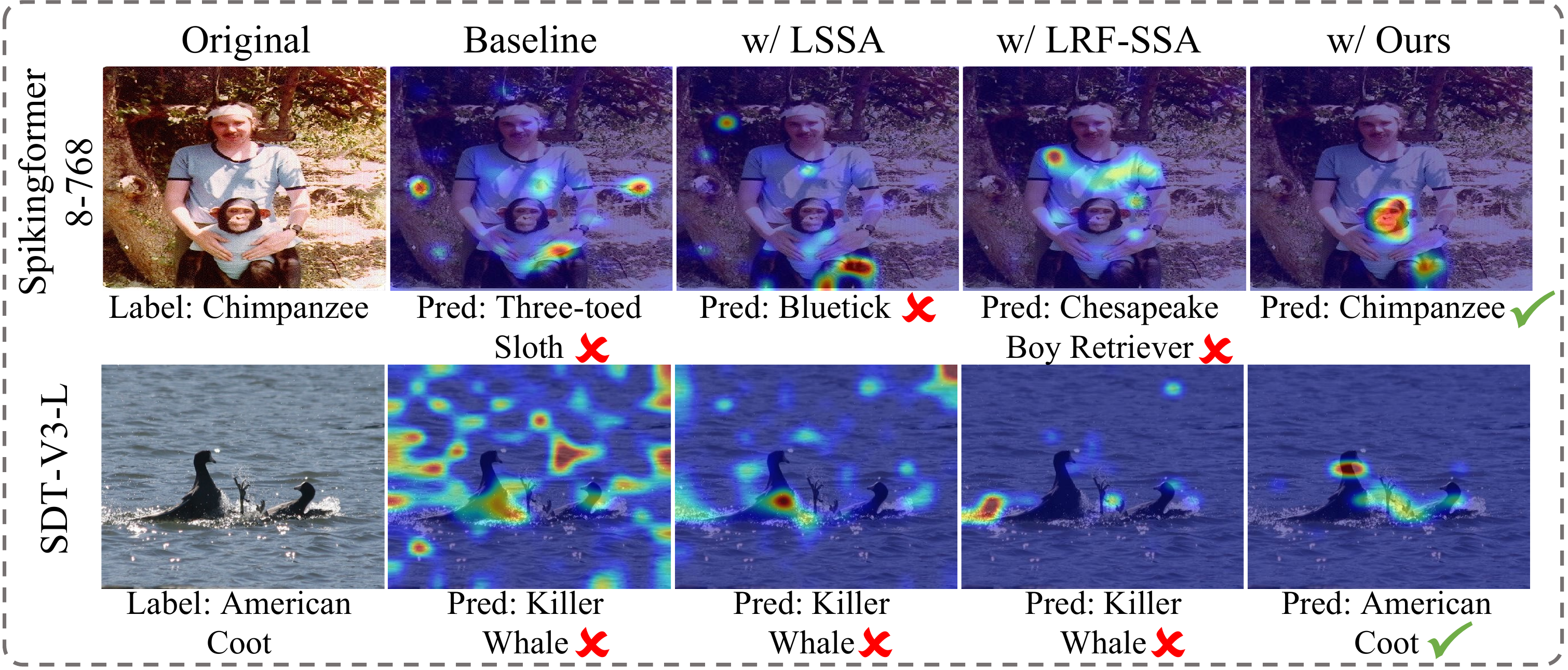}
  \caption{Comparative visualization on ImageNet-1K using Spikingformer (ViT-like plain architecture) and SDT-V3 (multi-stage hierarchical architecture) with different locality-enhanced SSA variants.}
  \label{fig:heatmap}
\end{figure}

To answer \textbf{RQ3}, Figure~\ref{fig:exp_SCLA_BCP_compare_variants} shows that our method yields lower MAD at every deployed layer across all three backbones.
The observed reduction is consistent with the design of SCLA-BCP, in which attention is restricted to spatially contiguous regions, thereby bounding its spatial interaction range.
We further visualize the learned spatial interaction patterns in Figure~\ref{fig:heatmap}. 
Compared with LSSA and LRF-SSA, our proposed approach produces more concentrated responses on foreground objects while suppressing scattered activations in irrelevant regions, suggesting that our method learns more effective spatial representations. Additional visualization results are provided in Appendix~G.

\subsection{Ablation Study}
To analyze the components and design choices of SCLA-BCP, we perform ablation studies on QKFormer (multi-stage hierarchical architecture) and Spikingformer (ViT-like plain architecture) using CIFAR-100 ($T=4$) and CIFAR10-DVS ($T=16$). The training configurations and deployment strategies are consistent with the main experiments.

\noindent\textbf{Contributions of SCLA and BCP.}
Table~\ref{tab:component_ablation} decomposes the two components. SCLA alone consistently improves accuracy over baselines (up to 1.20\% on CIFAR10-DVS), validating the benefit of confining attention to local contiguous regions, whereas BCP alone provides only marginal gains on static data (0.05\%/0.07\% on CIFAR-100). Combining both components further improves performance, indicating that BCP does not serve as an additional pathway but compensates for the restricted cross-region information exchange caused by SCLA partitioning.
\begin{table}[htbp]
\centering
\footnotesize
\setlength{\tabcolsep}{3pt}
\renewcommand{\arraystretch}{0.90}
\begin{tabular}{llcccc}
\toprule
\multirow{2}{*}{\textbf{Backbone}}
& \multirow{2}{*}{\textbf{Dataset}}
& \multirow{2}{*}{\textbf{Baseline}}
& \textbf{SCLA}
& \textbf{BCP}
& \multirow{2}{*}{\textbf{Ours}} \\
& &
& \textbf{only}
& \textbf{only}
& \\
\midrule
\multirow{2}{*}{QKFormer}
& CIFAR-100
& 80.84 & 81.05 & 80.89 & \textbf{81.21} \\
& CIFAR10-DVS
& 81.10 & 82.20 & 81.90 & \textbf{83.50} \\
\midrule
\multirow{2}{*}{Spikingformer}
& CIFAR-100
& 80.37 & 80.61 & 80.44 & \textbf{80.91} \\
& CIFAR10-DVS
& 81.40 & 82.60 & 81.90 & \textbf{83.60} \\
\bottomrule
\end{tabular}

\caption{Component ablation results on CIFAR-100 ($T=4$) and CIFAR10-DVS ($T=16$). The SCLA-only setting removes the BCP pathway, while the BCP-only setting replaces SCLA with the original SSA.}
\label{tab:component_ablation}
\end{table}

\noindent\textbf{Comparison with shifted-window designs.}
We examine whether shifted-window attention (SWA) can replace BCP by removing the BCP pathway and applying the SCLA region size $w$ to SWA~\citep{liu2021swin}. As shown in Table~\ref{tab:shift_ablation}, SWA without relative position encoding (RPE)~\citep{wu2021rethinking} performs comparably to SCLA-only, indicating limited benefit from window shifting alone. 
Standard SWA still underperforms SCLA-BCP by 0.44\% on average. 
This result is expected, as conventional positional encodings are difficult to effectively exploit in SNNs \citep{lv2024advancing, lv2026toward}, and the introduction of floating-point RPE compromises the spike-driven computation property.

\begin{table}[htbp]
\centering
\footnotesize
\setlength{\tabcolsep}{3pt}
\begin{tabular}{@{}llcccc@{}}
\toprule
\multirow{2}{*}{\textbf{Backbone}}
& \multirow{2}{*}{\textbf{Dataset}}
& \textbf{SCLA}
& \textbf{SWA}
& \textbf{SWA}
& \multirow{2}{*}{\textbf{Ours}} \\
& & \textbf{only}
& \textbf{w/o RPE}
& \textbf{w/ RPE}
& \\
\midrule
\multirow{2}{*}{QKFormer}
& CIFAR-100
& 81.05 & 81.02 & 81.03 & \textbf{81.21} \\
& CIFAR10-DVS
& 82.20 & 82.20 & 82.90 & \textbf{83.50} \\
\midrule
\multirow{2}{*}{Spikingformer}
& CIFAR-100
& 80.61 & 80.70 & 80.73 & \textbf{80.91} \\
& CIFAR10-DVS
& 82.60 & 82.30 & 82.80 & \textbf{83.60} \\
\bottomrule
\end{tabular}
\caption{Comparison of different cross-region interaction mechanisms for SCLA on CIFAR-100 ($T=4$) and CIFAR10-DVS ($T=16$).}
\label{tab:shift_ablation}
\end{table}

\section{Conclusion}

In this work, we investigate the spatial locality issue in SSA and reveal that existing locality-enhanced SSA methods struggle to establish localized token interactions due to a computational--spatial locality discrepancy and uniform locality deployment across different architectures.
We propose SCLA-BCP, where SCLA partitions feature maps into local contiguous regions and restricts attention computation within each region, while BCP facilitates information exchange across region boundaries. 
We further develop a hierarchical deployment strategy to effectively apply SCLA-BCP to different architectures.
Extensive experiments across diverse vision tasks demonstrate that our method consistently improves the performance of Spiking Transformers with limited overhead. 
Additional qualitative and quantitative analyses, including visualization and ablation studies, validate the effectiveness of our method.

\bibliographystyle{plainnat}
\bibliography{references}

\end{document}